\documentclass[conference]{IEEEtran}
\IEEEoverridecommandlockouts
\usepackage{cite}
\usepackage{amsmath,amssymb,amsfonts}
\usepackage{algorithmic}
\usepackage{graphicx}
\usepackage{textcomp}
\usepackage{xcolor}
\usepackage{tikz}

\usepackage{pgfplots}
\usepackage{pgfplotstable}
\pgfplotsset{compat=1.7}
\usepgfplotslibrary{groupplots}
\usepackage{subcaption}
\usepackage{soul} 
\sethlcolor{orange}
\usepackage[normalem]{ulem} 

\usetikzlibrary{external}

\usepackage{amsmath}
\DeclareMathOperator{\pred}{Pred}
\DeclareMathOperator{\layerv}{Layer^v}
\DeclareMathOperator{\layerf}{Layer^f}

\def\BibTeX{{\rm B\kern-.05em{\sc i\kern-.025em b}\kern-.08em
    T\kern-.1667em\lower.7ex\hbox{E}\kern-.125emX}}
    
\newcommand\scalemath[2]{\scalebox{#1}{\mbox{\ensuremath{\displaystyle #2}}}}    
    
\begin{document}

\title{State Estimation in Electric Power Systems Leveraging Graph Neural Networks \\

\thanks{This paper has received funding from the European Union’s Horizon 2020 research and innovation programme under Grant Agreement number 856967.}
}

\author{\IEEEauthorblockN{Ognjen Kundacina}
\IEEEauthorblockA{\textit{The Institute for Artificial Intelligence } \\
\textit{Research and Development of Serbia }\\
Novi Sad, Serbia \\
ognjen.kundacina@ivi.ac.rs}
\and
\IEEEauthorblockN{Mirsad Cosovic}
\IEEEauthorblockA{\textit{Faculty of Electrical
Engineering} \\
\textit{University of Sarajevo}\\
Sarajevo, Bosnia and Herzegovina \\
mcosovic@etf.unsa.ba}
\and
\IEEEauthorblockN{Dejan Vukobratovic}
\IEEEauthorblockA{\textit{Faculty of Technical Sciences} \\
\textit{University of Novi Sad}\\
Novi Sad, Serbia \\
dejanv@uns.ac.rs}
}

\IEEEoverridecommandlockouts
\IEEEpubid{\makebox[\columnwidth]{978-1-6654-1211-7/22/\$31.00~\copyright2022 IEEE \hfill} \hspace{\columnsep}\makebox[\columnwidth]{ }}

\maketitle

\IEEEpubidadjcol

\begin{abstract}
The goal of the state estimation (SE) algorithm is to estimate complex bus voltages as state variables based on the available set of measurements in the power system. Because phasor measurement units (PMUs) are increasingly being used in transmission power systems, there is a need for a fast SE solver that can take advantage of high sampling rates of PMUs. This paper proposes training a graph neural network (GNN) to learn the estimates given the PMU voltage and current measurements as inputs, with the intent of obtaining fast and accurate predictions during the evaluation phase. GNN is trained using synthetic datasets, created by randomly sampling sets of measurements in the power system and labelling them with a solution obtained using a linear SE with PMUs solver. The presented results display the accuracy of GNN predictions in various test scenarios and tackle the sensitivity of the predictions to the missing input data.
\end{abstract}


\section{Introduction}
The state estimation (SE), which estimates the set of power system state variables based on the available set of measurements, is an essential tool used for the power system's monitoring and operation \cite{monticelli2000SE}. A fast state estimator is required to maximise the use of high sampling rates of phasor measurement units (PMUs). In this paper, we propose training a graph neural network (GNN) for a regression problem on a dataset of SE inputs and outputs, to provide fast and accurate predictions during the evaluation phase. One of the main benefits of using GNNs in power systems instead of common deep learning methods is the fact that the prediction model is not limited to the training and test examples of fixed power system topologies. GNNs exploit the graph structure of the input data \cite{pmlr-v70-gilmer17a}\cite{geomDeepLearning}, resulting in a lower number of learning parameters, reduced memory requirements, and incorporate the connectivity information into the learning process as well. Furthermore, the inference phase of the trained GNN model can be distributed, since the prediction of the state variable of a node requires only $K$-hop neighbourhood measurements. 

\section{Related Research and Contributions}
Several pieces of research suggest learning computationally heavy algorithms' outputs using a deep learning model, trained on the set of examples generated by the algorithm offline. In \cite{zhang2019} combination of recurrent and feed-forward neural networks is used to solve the power system SE problem, given the measurement data and the history of network voltages. An example of training a feed-forward neural network to initialise the network voltages for the Gauss-Newton distribution system SE solver is given in \cite{zamzam2019}. GNNs are beginning to be used for solving similar problems, like in \cite{donon2019graphneuralsolver}, where power flows in the system are predicted based on the power injection data labelled by the traditional power flow solver. In paper \cite{pagnier2021physicsinformed}, the authors propose a combined model and data-based approach in which GNNs are used for the power system parameter and state estimation. The model predicts power injections/consumptions in the nodes where voltages and phases are measured, whereas it doesn't include the branch measurements, as well as node measurement types other than the voltage in the calculation. In \cite{Yang2020RobustPU}, GNN was trained to provide state variable initialization to existing SE solver by propagating simulated or measured voltages through the graph to learn the voltage labels from the historical dataset. However, the proposed GNN also does not take into account measurement types other than node voltage measurements, but they are handled in the other parts of the algorithm. 

In this paper, the proposed model operates on factor-graph-like structures, which enables trivial inclusion and exclusion of any type of measurements on the power system's buses and branches, by adding or removing the corresponding nodes in the graph. The trained model is tested thoroughly in the various missing data scenarios which include communication errors in the delivery of isolated phasor data or failures of the complete PMUs, in which the power system is unobservable, and the reports on the prediction qualities are presented.

\section{Linear State Estimation with PMUs}
The SE algorithm estimates the values of the state variables $\mathbf{x}$ based on the knowledge of the network topology and parameters, and measured values obtained from the measurement devices spread across the power system. 

The power system network topology is described by the bus/branch model and can be represented using a graph $\mathcal{G} =(\mathcal{H},\mathcal{E})$, where the set of nodes $\mathcal{H} = \{1,\dots,n  \}$ represents the set of buses, while the set of edges $\mathcal{E} \subseteq \mathcal{H} \times \mathcal{H}$ represents the set of branches of the power network. The branches of the network are defined using the two-port $\pi$-model. More precisely, the branch $(i,j) \in \mathcal{E}$ between buses $\{i,j\} \in \mathcal{H}$ can be modelled using complex expressions:
\begin{equation}
  \begin{bmatrix}
    {I}_{ij} \\ {I}_{ji}
  \end{bmatrix} =
  \begin{bmatrix}
    \cfrac{1}{\tau_{ij}^2}(y_{ij} + y_{\text{s}ij}) & -\alpha_{ij}^*{y}_{ij}\\
    -\alpha_{ij}{y}_{ij} & {y}_{ij} + y_{\text{s}ij}
  \end{bmatrix}  
  \begin{bmatrix}
    {V}_{i} \\ {V}_{j}
  \end{bmatrix},
  \label{unified}
\end{equation} 
where the parameter $y_{ij} = g_{ij} + \text{j}b_{ij}$ represents the branch series admittance, half of the total branch shunt admittance (i.e., charging admittance) is given as $y_{\text{s}ij} = \text{j}b_{si}$. Further, the transformer complex ratio is defined as $\alpha_{ij} = (1/\tau_{ij})\text{e}^{-\text{j}\phi_{ij}}$, where $\tau_{ij}$ is the transformer tap ratio magnitude, while $\phi_{ij}$ is the transformer phase shift angle. It is important to remember that the transformer is always located at the bus $i$ of the branch described by \eqref{unified}. Using the branch model defined by \eqref{unified}, if $\tau_{ij} = 1$ and $\phi_{ij} = 0$ the system of equations describes a line. In-phase transformers are defined if $\phi_{ij} = 0$ and $y_{\text{s}ij} = 0$, while phase-shifting transformers are obtained if $y_{\text{s}ij} = 0$. The complex expressions ${I}_{ij}$ and ${I}_{ji}$ define branch currents from the bus $i$ to the bus $j$, and from the bus $j$ to the bus $i$, respectively. The complex bus voltages at buses $\{i,j\}$ are given as $V_i$ and $V_j$, respectively.  

PMUs measure complex bus voltages and complex branch currents. More precisely, phasor measurement provided by PMU is formed by a magnitude, equal to the root mean square value of the signal, and phase angle \cite[Sec.~5.6]{phadke}. The PMU placed at the bus measures bus voltage phasor and current phasors along all branches incident to the bus \cite{exposito}. Thus, the PMU outputs phasor measurements in polar coordinates. In addition, PMU outputs can be observed in the rectangular coordinates with real and imaginary parts of the bus voltage and branch current phasors. In that case, the vector of state variables $\mathbf{x}$ can be given in rectangular coordinates $\mathbf x \equiv[\mathbf{V}_\mathrm{re},\mathbf{V}_\mathrm{im}]^{\mathrm{T}}$, where we can observe real and imaginary components of bus voltages as state variables:   
\begin{equation}
   	\begin{aligned}
        \mathbf{V}_\mathrm{re}&=\big[\Re({V}_1),\dots,\Re({V}_n)\big]\\
	    \mathbf{V}_\mathrm{im}&=\big[\Im({V}_1),\dots,\Im({V}_n)\big].     
   	\end{aligned}
   	\label{rect_coord}
\end{equation} 

Using rectangular coordinates, we obtain the linear system of equations defined by voltage and current measurements obtained from PMUs. The measurement functions corresponding to the bus voltage phasor measurement on the bus $i \in \mathcal{H}$ are simply equal to: 
\begin{equation}
    \begin{aligned}
        f_{\Re\{V_i\}}(\cdot) = \Re\{V_i\}\\
        f_{\Im\{V_i\}}(\cdot) = \Im\{V_i\}.
    \end{aligned}    
\end{equation}
According to the unified branch model \eqref{unified}, functions corresponding to the branch current phasor measurement vary depending on where the PMU is located. If PMU is placed at the bus $i$ where the transformer is located, functions are given as:
\begin{equation}
\scalemath{0.83}{
    \begin{aligned}
        f_{\Re\{I_{ij}\}}(\cdot) &= q \Re\{V_{i}\} - w \Im\{V_{i}\} - (r-t) \Re\{V_{j}\} + (u+p) \Im\{V_{j}\} \\
        f_{\Im\{I_{ij}\}}(\cdot) &= w \Re\{V_{i}\} + q \Im\{V_{i}\} - (u+p) \Re\{V_{j}\} - (r-t) \Im\{V_{j}\},
    \end{aligned}}
\end{equation}
where $q =$ $g_{ij}/\tau_{ij}^2$, $w =$ $(b_{ij} + b_{si})/\tau_{ij}^2$, $r =$ $(g_{ij}/\tau_{ij})$ $\cos\phi_{ij}$, $t =$ $(b_{ij}/\tau_{ij})$ $\sin\phi_{ij}$, $u =$ $(b_{ij}/\tau_{ij})$ $\cos\phi_{ij}$, $p =$ $(g_{ij}/\tau_{ij})$ $\sin\phi_{ij}$. In the case where PMU is installed at the bus $j$, at the opposite side of the transformer, measurement functions are: 
\begin{equation}
\scalemath{0.83}{
    \begin{aligned}
        f_{\Re\{I_{ji}\}}(\cdot) &= z \Re\{V_{j}\} - e \Im\{V_{j}\} - (r+t) \Re\{V_{i}\} + (u-p) \Im\{V_{i}\} \\
        f_{\Im\{I_{ji}\}}(\cdot) &= e \Re\{V_{j}\} + z \Im\{V_{j}\} - (u-p) \Re\{V_{i}\} - (r+t) \Im\{V_{i}\},
    \end{aligned}}
\end{equation}
where $z = g_{ij}$ and $e =$ $b_{ij} + b_{si}$. To recall, the presented model represents the system of linear equations, where the solution can be found by solving the linear weighted least-squares (WLS) problem: 
\begin{equation}
    \left(\mathbf H^{T} \mathbf R^{-1} \mathbf H \right) \mathbf x =
		\mathbf H^{T} \mathbf R^{-1} \mathbf z,    
	\label{SE_system_of_lin_eq}
\end{equation}
where the Jacobian matrix $\mathbf {H} \in \mathbb {R}^{k \times 2n}$ is defined according to measurement functions, $k$ is the total number of linear equations, the covariance matrix is given as $\mathbf {R} \in \mathbb {R}^{k \times k}$, and the vector $\mathbf z \in \mathbb {R}^{k}$ contains measurement values given in rectangular coordinate system. 


\section{Graph Neural Network based State Estimation}

GNNs are a suitable tool for learning over graph-structured data, which is being processed by following a recursive neighbourhood aggregation scheme, also known as message passing procedure \cite{pmlr-v70-gilmer17a}. This results in an $s$-dimensional vector embedding $\mathbf h \in \mathbb {R}^{s}$ of each node, which captures the information about the node's position in the graph, as well as it's own and the input features of the neighboring nodes. The GNN layer, which implements one iteration of the recursive neighbourhood aggregation consists of several functions, that can be represented using a trainable set of parameters, usually in form of the feed-forward neural networks. Those include the message function between two node embeddings, the aggregation function which defines in which way are incoming messages combined and the update function which updates the node embedding based on the aggregated messages and the current node embedding value. The recursive neighbourhood aggregation scheme is repeated a predefined number of times $K$, also known as the number of GNN layers, where the initial node embedding values are equal to the $l$-dimensional node input features, linearly transformed to the initial node embedding $\mathbf h^0 \in \mathbb {R}^{s}$. The output of this process are final node embeddings which can be used for the classification or regression over the nodes, edges, or the whole graph, or can be used directly for the unsupervised node or edge analysis of the graph. In the case of supervised learning over the nodes, the final embeddings are passed through the additional nonlinear function, creating the outputs that represent the predictions of the GNN model for the set of inputs fed into the nodes and their neighbors. The GNN model is trained by backpropagation of the loss function between the labels and the predictions over the whole computational graph.

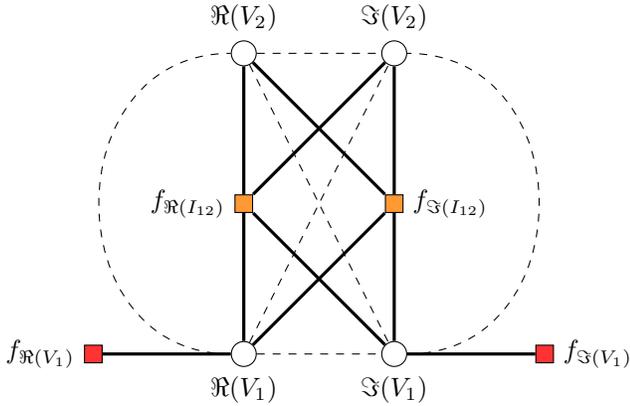
\begin{figure}[htbp]
    \centering
    \begin{tikzpicture} [scale=1.0, transform shape]
        \tikzset{
            varNode/.style={circle,minimum size=2mm,fill=white,draw=black},
            factorVoltage/.style={draw=black,fill=red!80, minimum size=2mm},
            factorCurrent/.style={draw=black,fill=orange!80, minimum size=2mm},
            edge/.style={very thick,black},
            edge2/.style={dashed,black}}
        \begin{scope}[local bounding box=graph]
            \node[factorVoltage, label=left:$f_{\Re({V}_1)}$] (f1) at (-3, 1 * 2) {};
            \node[factorVoltage, label=right:$f_{\Im({V}_1)}$] (f4) at (3, 1 * 2) {};
            \node[varNode, label=below:$\Re({V}_1)$] (v1) at (-1, 1 * 2) {};
            \node[varNode, label=below:$\Im({V}_1)$] (v3) at (1, 1 * 2) {};
            \node[factorCurrent, label=left:$f_{\Re({I}_{12})}$] (f2) at (-1, 2 * 2) {};
            \node[factorCurrent, label=right:$f_{\Im({I}_{12})}$] (f5) at (1, 2 * 2) {};
            \node[varNode, label=above:$\Re({V}_2)$] (v2) at (-1, 3 * 2) {};
            \node[varNode, label=above:$\Im({V}_2)$] (v4) at (1, 3 * 2) {};
            
            \draw[edge] (f1) -- (v1);
            \draw[edge] (f4) -- (v3);
            \draw[edge] (f2) -- (v1);
            \draw[edge] (f5) -- (v3);
            \draw[edge] (f5) -- (v1);
            \draw[edge] (f2) -- (v3);
            \draw[edge] (f2) -- (v2);
            \draw[edge] (f2) -- (v4);
            \draw[edge] (f5) -- (v2);
            \draw[edge] (f5) -- (v4);
            \draw[edge2] (v1) to [out=180,in=180,looseness=1.5] (v2);
            \draw[edge2] (v1) -- (v3);
            \draw[edge2] (v1) -- (v4);
            \draw[edge2] (v2) -- (v3);
            \draw[edge2] (v2) -- (v4);
            \draw[edge2] (v3) to [out=0,in=0,looseness=1.5] (v4);
            
        \end{scope}
    \end{tikzpicture}
    \caption{Example of the factor graph (full-line edges) and the augmented factor graph for a two-bus power system. Variable nodes are depicted as circles, and factor nodes as squares, colored differently to distinguish between measurement types.}
    \label{toyFactorGraph}
    
\end{figure}

Inspired by the recent work \cite{cosovic2019bpse} in which the power system SE is modelled as a probabilistic graphical model, the initial version of the graph over which GNN operates has a power system's factor graph topology, which is a bipartite graph consisted of the factor and variable nodes, and the edges between them. Variable nodes, two per each power system bus, learn the $s$-dimensional representation of the state variables $\mathbf{x}$, i.e. real and imaginary parts of the bus voltages, $\Re({V}_i)$ and $\Im({V}_i)$, defined in \eqref{rect_coord}. Factor nodes, two per each measurement phasor, serve as inputs for the measurement values and variances, also given in rectangular coordinates, and whose embedded values are sent to variable nodes via GNN message passing. Feature augmentation using one-hot index encoding is performed for variable nodes only, to help the GNN model to represent the neighbourhood structure of a node better, since variable nodes have no additional input features. Nodes connected by full lines in Fig.~\ref{toyFactorGraph} represent a simple example of the factor graph for a two-bus power system, with a PMU on the first bus, containing one voltage and one current phasor measurement. Compared to the approaches like \cite{pagnier2021physicsinformed}, in which GNN nodes correspond to state variables only, we find factor-graph-like GNN topology convenient for incorporating measurements in the GNN, because factor nodes can be added or removed from any place in the graph, using which one can simulate inclusion of various types and quantities of measurements both on power system buses and branches.

We extend this approach by augmenting the graph topology by connecting the variable nodes in the 2-hop neighbourhood, following the idea that the graph should stay connected even in a case of simulating measurement loss by removing the factor nodes, enabling the messages to still be propagated in the whole $K$-hop neighbourhood of the variable node. In other words, a factor node corresponding to a branch current measurement can be removed, while still preserving the physical connection that exists between the power system buses. This requires adding an additional set of trainable parameters for variable-to-variable message function. We will still use the terms factor and variable nodes, although the second version graph over which the GNN operates, displayed in Fig.~\ref{toyFactorGraph} with additional edges depicted with dashed lines is not a factor graph, since it is no longer bipartite.

Since our GNN operates on a heterogeneous graph, in this work we employ two different types of GNN layers, $\layerf(\cdot|\theta^{\layerf}): \mathbb {R}^{\textrm{deg}(f)+1} \mapsto \mathbb {R}^{s}$ for aggregation in factor nodes $f$, and $\layerv(\cdot|\theta^{\layerv}): \mathbb {R}^{\textrm{deg}(v)+1} \mapsto \mathbb {R}^{s}$ for variable nodes $v$, so that their message, aggregation, and update functions are learned using a separate set of trainable parameters, denoted all together as  $\theta^{\layerf}$ and $\theta^{\layerv}$. Additionally, we use a different set of parameters for variable-to-variable and factor-to-variable node messages in the $\layerv(\cdot|\theta^{\layerv})$ layer. In both GNN layers, we used two-layer feed-forward neural networks as message functions, gated recurrent units as update functions and the attention mechanism in the aggregation function \cite{velickovic2018graph}, using which the importance factor of each neighbour is learned. Furthermore, we apply a two-layer neural network $\pred(\cdot|\theta^{\pred}): \mathbb {R}^{s} \mapsto \mathbb {R}$ on top of the final node embeddings $\mathbf h^K$ of variable nodes only, to create the state variable predictions $\mathbf{x^{pred}}$. For the factor and variable nodes with indices $f$ and $v$, neighbourhood aggregation, and state variable prediction can be described as: 
\begin{equation}
    \begin{gathered}
        {h_v}^k = \layerv(\{{h_i}^{k-1} | i \in \{v \mathop{\cup}  \mathcal{N}_v\}\} | \theta^{\layerv})\\
        {h_f}^k = \layerf(\{{h_i}^{k-1} | i \in \{f \mathop{\cup}  \mathcal{N}_f\}\} | \theta^{\layerf})\\
        {x_v}^{pred} = \pred({h_v}^K|\theta^{\pred})\\
        k \in \{1,\dots,K\}.
    \end{gathered}
    \label{embeddingds_and_predictions}
\end{equation}

All of the GNN trainable parameters $\theta$ are updated by applying gradient descent (i.e. backpropagation) to a loss function calculated over the whole mini-batch of graphs, as a mean squared difference between the state variable predictions and labels $\mathbf{x^{label}}$:
\begin{equation} \label{loss_function}
    \begin{gathered}
        L(\theta) = \frac{1}{2nB} \sum_{i=1}^{2nB}({{x_i}^{pred}} - {{x_i}^{label}})^2 \\
        \theta = \{\theta^{\layerf} \mathop{\cup} \theta^{\layerv} \mathop{\cup} \theta^{\pred}\},
    \end{gathered}
\end{equation}
where $2n$ is the total number of variable nodes in a graph, and $B$ is the number of graphs in the mini-batch.

The inference process using the trained model can be computationally and geographically distributed, as long as all of the measurements within the $K$-hop neighbourhood in the augmented factor graph are fed into the computational module that generates the predictions. For arbitrary $K$, PMUs required for the inference will be physically located within the $\lceil K/2 \rceil$-hop neighbourhood of the power system bus.

\section{Numerical results}
In this section, we describe the training setup which is used for both GNN models described in the previous section and assess the prediction quality of the trained models in various test scenarios. Training, validation, and test sets are obtained using the WLS solutions of the system described in (\ref{SE_system_of_lin_eq}) for various measurement samples. The number and the positions on PMUs are fixed and determined using the optimal PMU placement algorithm \cite{optimalPMUPlacement}, which finds the smallest set of PMU measurements that make the system observable. The algorithm has resulted in a total of 50 measurement phasors, 10 of which are the voltage phasors and the rest are the current phasors. The IEEE 30-bus test case is the foundation for all of the datasets we used in our experiments, which is enriched by measurements, obtained by adding Gaussian noise to the exact power flow solutions, with each power flow calculation executed with different load profile.

The training set consists of 10000 samples, while validation and test set both contain 100 samples. The GNN model for factor graphs, as well as the model for augmented factor graphs, were both trained on 100 epochs, which was sufficient to reach convergence considering both training and validation loss functions. 
We used the IGNNITION framework \cite{pujolperich2021ignnition} for building and utilising GNN models, with the hyperparameters presented in Table~\ref{tbl_hyperparameters}, the first three of which were obtained with the grid search hyperparameter optimization. 

\begin{table}[!t]
\caption{List of GNN hyperparameters.}
\label{tbl_hyperparameters}
\begin{center}
    \begin{tabular}{ | l | c | }
        \hline
        \textbf{Hyperparameters} & \textbf{Values} \\ 
        \hline
        Node embedding size $s$ & $64$ \\
        \hline
        Learning rate & $4\times 10^{-4}$ \\
        \hline
        Minibatch size $B$ & $32$ \\
        \hline
        Number of GNN layers $K$ & $4$ \\
        \hline
        Activation functions & ReLU \\
        \hline
        Gradient clipping value & $0.5$ \\
        \hline
        Optimizer & Adam \\
        \hline
        Batch normalization type & Mean \\
        \hline
    \end{tabular}
\end{center}
\vspace{-6mm}
\end{table}

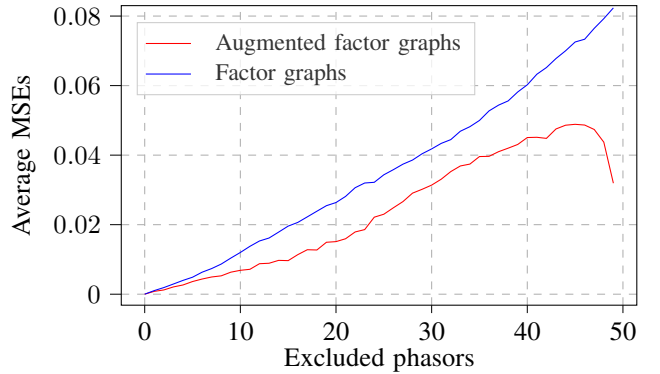
\begin{figure}[htbp]
    \centering
    \pgfplotstableread[col sep = comma,]{data/test_result_columns.csv}\testResultColumns
    \begin{tikzpicture}
    
    \begin{axis}[
    yscale=0.70,
    legend cell align={left},
    legend columns=1,
    legend style={
      fill opacity=0.8,
      font=\small,
      draw opacity=1,
      text opacity=1,
      at={(0.03,1.35)},
      anchor=north west,
      draw=white!80!black
    },
    tick align=outside,
    tick pos=left,
    x grid style={white!69.0196078431373!black},
    xlabel={Excluded phasors},
    xmin=-2.45, xmax=51.45,
    xtick style={color=black},
    y grid style={white!69.0196078431373!black},
    ylabel={Average MSEs},
    ymin=-0.00315987188222194, ymax=0.083,
    ytick style={color=black},
        yticklabel style={
        /pgf/number format/fixed,
        /pgf/number format/precision=5
    },
    scaled y ticks=false,
    xticklabels={0, 0,10,20,30,40,50},
    xmajorgrids=true,
    ymajorgrids=true,
    grid style=dashed
    ]
    \addplot [red]
    table [x expr=\coordindex, y={averageMSEs}]{\testResultColumns};
    \addlegendentry{Augmented factor graphs}
    \addplot [blue]
    table [x expr=\coordindex, y={averageMSEsRegularDataset}]{\testResultColumns};
    \addlegendentry{Factor graphs}
    
    \end{axis}
    
    \end{tikzpicture}
    \caption{Average MSEs of test sets created by randomly excluding measurement phasors.}
    \label{msesForBothModels}
\vspace{-4mm}    
\end{figure}

To assess the quality of the proposed GNN model in unobservable areas, we test the trained models by excluding various numbers of measurement phasors from the previously used test samples, making the system of equations that describes the SE problem underdetermined. Excluding the measurement phasor from the test sample is realised by removing its real and imaginary parts from the input data, which is equivalent to removing two factor nodes from the graph on which the GNN operates. Using the previously used 100-sample test set we create 49 additional test sets, by removing a number in a range of $[1, 49]$ measurement phasors randomly from each sample, while preserving the same labels obtained as SE solutions of the system with all of the measurements present. Fig.~\ref{msesForBothModels} summarises the quality of predictions of the both GNN models on the mentioned test sets. For the test set with no measurements excluded, the average mean square error (MSE) between the predictions and the labels for GNN models for factor graphs and augmented factor graphs equals $1.056\times 10^{-5}$ and $1.051\times 10^{-5}$, respectively, while for the test set with 49 measurement phasors excluded, the corresponding values are $8.2235\times 10^{-2}$ and $3.193\times 10^{-2}$. Both models displayed very low MSE on test examples with no excluded measurements, but the second model performed better on test sets with excluded measurements. This confirms the assumption that additional connections between variable nodes in the graph will preserve information exchange when factor nodes are removed. Therefore, we will analyse the second model in the detail in the rest of the paper.

\subsection{Analysis of results for the GNN model operating on augmented factor graphs}

\begin{figure}[htbp]
    \centering
    \pgfplotstableread[col sep = comma,]{data/PredictionsPerNode0Excluded.csv}\CSVPredVSObservedZeroExcludedEdgesVarNodes
    
        \begin{tikzpicture}
        
        \begin{groupplot}[group style={group size=1 by 2}, ]
        \nextgroupplot[
        yscale=0.86,
        legend cell align={left},
        legend columns=1,
        legend style={
          fill opacity=0.8,
          font=\small,
          draw opacity=1,
          text opacity=1,
          at={(0.5,0.05)},
          anchor=south,
          draw=white!80!black
        },
        scaled x ticks=manual:{}{\pgfmathparse{#1}},
        tick align=outside,
        tick pos=left,
        x grid style={white!69.0196078431373!black},
        xmin=-1.45, xmax=30.45,
        xtick style={color=black},
        xticklabels={},
        y grid style={white!69.0196078431373!black},
        ylabel={\(\displaystyle \Re({V}_i)\)},
        ymin=0.749538275942936, ymax=1.13792501106016,
        ytick style={color=black},
        xmajorgrids=true,
        ymajorgrids=true,
        grid style=dashed
        ]
        \addplot [red, mark=square*, mark options={scale=0.5}, mark repeat=2,mark phase=1]
        table [x expr=\coordindex, y={predictionsRE}]{\CSVPredVSObservedZeroExcludedEdgesVarNodes};
        \addlegendentry{Predictions}
        
        \addplot [blue, mark=square*, mark options={scale=0.5}, mark repeat=2,mark phase=2]
        table [x expr=\coordindex, y={true_valuesRE}]{\CSVPredVSObservedZeroExcludedEdgesVarNodes};
        \addlegendentry{Labels}
        
        \addplot [black, forget plot]
        table [x expr=\coordindex, y={max_voltagesRE}]{\CSVPredVSObservedZeroExcludedEdgesVarNodes};
        
        \addplot [black]
        table [x expr=\coordindex, y={min_voltagesRE}]{\CSVPredVSObservedZeroExcludedEdgesVarNodes};
        \addlegendentry{Bounds}

        \nextgroupplot[
        yscale=0.86,
        tick align=outside,
        tick pos=left,
        x grid style={white!69.0196078431373!black},
        xlabel={Bus index ($i$)},
        xmin=-1.45, xmax=30.45,
        xtick style={color=black},
        y grid style={white!69.0196078431373!black},
        ylabel={\(\displaystyle \Im({V}_i)\)},
        ymin=-0.537296454651957, ymax=0.0692334804622149,
        ytick style={color=black},
        xmajorgrids=true,
        ymajorgrids=true,
        grid style=dashed
        ]
        
        \addplot [red, mark=square*, mark options={scale=0.5}, mark repeat=2,mark phase=1]
        table [x expr=\coordindex, y={predictionsIM}]{\CSVPredVSObservedZeroExcludedEdgesVarNodes};
        
        \addplot [blue, mark=square*, mark options={scale=0.5}, mark repeat=2,mark phase=2]
        table [x expr=\coordindex, y={true_valuesIM}]{\CSVPredVSObservedZeroExcludedEdgesVarNodes};
        
        \addplot [black, forget plot]
        table [x expr=\coordindex, y={max_voltagesIM}]{\CSVPredVSObservedZeroExcludedEdgesVarNodes};
        
        \addplot [black]
        table [x expr=\coordindex, y={min_voltagesIM}]{\CSVPredVSObservedZeroExcludedEdgesVarNodes};
        
        \end{groupplot}
        
        \end{tikzpicture}
    \caption{Predictions and labels for one test example with no measurement phasors removed.}
    \label{PredictionsPerNode0Excluded}
\end{figure}
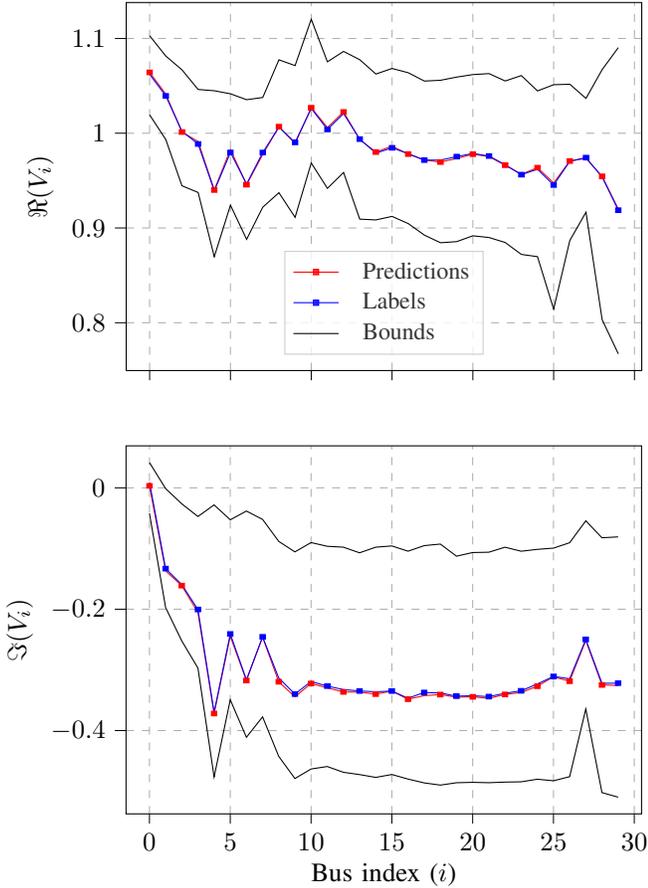

Average MSEs and Pearson’s Correlations between the prediction and the labels over the whole test sets with various numbers of excluded measurement phasors are presented in Table~\ref{tbl_excluding_measurements_1}. To further depict the quality of the trained model, for one of the test samples, in Fig.~\ref{PredictionsPerNode0Excluded} we present the predictions and the labels for each of the variable nodes, where the upper plot displays the results for the real parts of complex node voltages, while the lower plot displays the imaginary parts. Fig.~\ref{PredictionsPerNode0Excluded} also displays the minimal and the maximal value of the state variables for each node, looking over every sample in the training, validation, and test set. Presented results indicate that GNNs can be used as accurate SE approximators when trained on the representative dataset. 

\begin{table}[htbp]
\caption{Average MSEs and Pearson’s Correlations for test sets with various numbers of randomly excluding measurement phasors.}
\begin{center}
    \begin{tabular}{ | c | c | c | }
        \hline
        \textbf{Excluded phasors} & \textbf{Pearson’s Correlation} & \textbf{MSE} \\
        \hline
        0 & 0.9999948 & $1.0512\times 10^{-5}$ \\
        \hline
        2 & 0.9986435 & $1.2259\times 10^{-5}$ \\
        \hline
        10 & 0.9947819 & $6.8697\times 10^{-3}$ \\
        \hline
        25 & 0.9880212 & $2.3002\times 10^{-2}$ \\
        \hline
        49 & 0.9697979 & $3.1927\times 10^{-2}$ \\
        \hline
    \end{tabular}
\label{tbl_excluding_measurements_1}
\end{center}
\end{table}

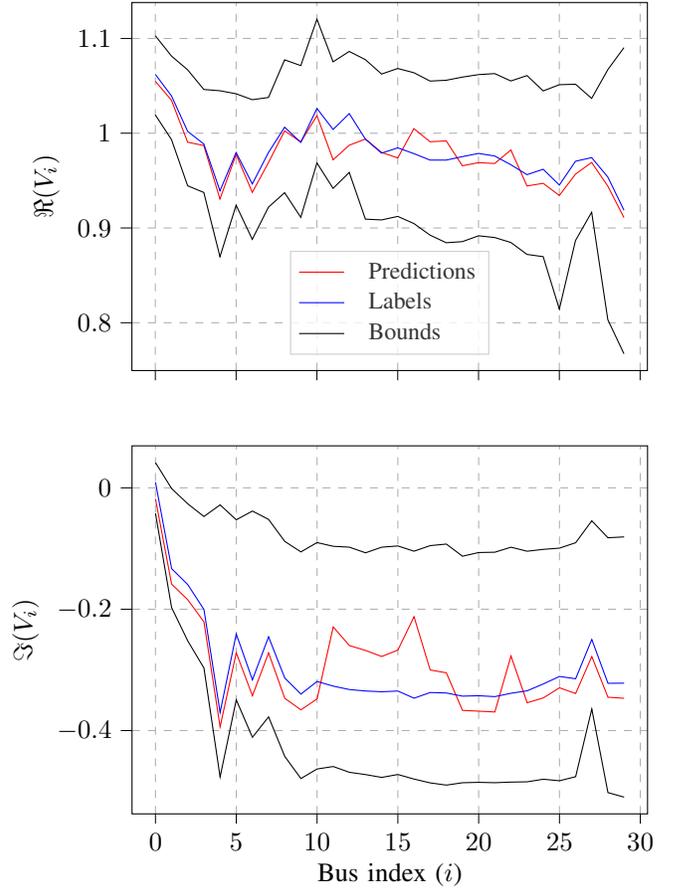
\begin{figure}[htbp]
    \centering
    \pgfplotstableread[col sep = comma,]{data/PredictionsPerNode2Excluded.csv}\CSVPredVSObservedTwoExcludedEdgesVarNodes
        \begin{tikzpicture}
        
        \begin{groupplot}[group style={group size=1 by 2}, ]
        \nextgroupplot[
        yscale=0.86,
        legend cell align={left},
        legend columns=1,
        legend style={
          fill opacity=0.8,
          font=\small,
          draw opacity=1,
          text opacity=1,
          at={(0.5,0.05)},
          anchor=south,
          draw=white!80!black
        },
        scaled x ticks=manual:{}{\pgfmathparse{#1}},
        tick align=outside,
        tick pos=left,
        x grid style={white!69.0196078431373!black},
        xmin=-1.45, xmax=30.45,
        xtick style={color=black},
        xticklabels={},
        y grid style={white!69.0196078431373!black},
        ylabel={\(\displaystyle \Re({V}_i)\)},
        ymin=0.749538275942936, ymax=1.13792501106016,
        ytick style={color=black},
        xmajorgrids=true,
        ymajorgrids=true,
        grid style=dashed
        ]
        \addplot [red]
        table [x expr=\coordindex, y={predictionsRE}]{\CSVPredVSObservedTwoExcludedEdgesVarNodes};
        \addlegendentry{Predictions}
        
        \addplot [blue]
        table [x expr=\coordindex, y={true_valuesRE}]{\CSVPredVSObservedTwoExcludedEdgesVarNodes};
        \addlegendentry{Labels}
        
        \addplot [black, forget plot]
        table [x expr=\coordindex, y={max_voltagesRE}]{\CSVPredVSObservedTwoExcludedEdgesVarNodes};
        
        \addplot [black]
        table [x expr=\coordindex, y={min_voltagesRE}]{\CSVPredVSObservedTwoExcludedEdgesVarNodes};
        \addlegendentry{Bounds}

        \nextgroupplot[
        yscale=0.86,
        tick align=outside,
        tick pos=left,
        x grid style={white!69.0196078431373!black},
        xlabel={Bus index ($i$)},
        xmin=-1.45, xmax=30.45,
        xtick style={color=black},
        y grid style={white!69.0196078431373!black},
        ylabel={\(\displaystyle \Im({V}_i)\)},
        ymin=-0.537296454651957, ymax=0.0692334804622149,
        ytick style={color=black},
        xmajorgrids=true,
        ymajorgrids=true,
        grid style=dashed
        ]
        
        \addplot [red]
        table [x expr=\coordindex, y={predictionsIM}]{\CSVPredVSObservedTwoExcludedEdgesVarNodes};
        
        \addplot [blue]
        table [x expr=\coordindex, y={true_valuesIM}]{\CSVPredVSObservedTwoExcludedEdgesVarNodes};
        
        \addplot [black, forget plot]
        table [x expr=\coordindex, y={max_voltagesIM}]{\CSVPredVSObservedTwoExcludedEdgesVarNodes};
        
        \addplot [black]
        table [x expr=\coordindex, y={min_voltagesIM}]{\CSVPredVSObservedTwoExcludedEdgesVarNodes};
        
        \end{groupplot}
        
        \end{tikzpicture}
    \caption{Predictions and labels for one test example with two measurement phasors removed.}
    \label{PredictionsPerNode2Excluded}
\end{figure}

\begin{figure}[htbp]
    \centering
    \pgfplotstableread[col sep = comma,]{data/PredictionsPerNode49Excluded.csv}\CSVPredVSObservedAllExcludedEdgesVarNodes
        \begin{tikzpicture}
        
        \begin{groupplot}[group style={group size=1 by 2}, ]
        \nextgroupplot[
        yscale=0.833,
        legend cell align={left},
        legend columns=1,
        legend style={
          fill opacity=0.8,
          font=\small,
          draw opacity=1,
          text opacity=1,
          at={(0.5,0.03)},
          anchor=south,
          draw=white!80!black
        },
        scaled x ticks=manual:{}{\pgfmathparse{#1}},
        tick align=outside,
        tick pos=left,
        x grid style={white!69.0196078431373!black},
        xmin=-1.45, xmax=30.45,
        xtick style={color=black},
        xticklabels={},
        y grid style={white!69.0196078431373!black},
        ylabel={\(\displaystyle \Re({V}_i)\)},
        ymin=0.749538275942936, ymax=1.13792501106016,
        ytick style={color=black},
        xmajorgrids=true,
        ymajorgrids=true,
        grid style=dashed
        ]
        \addplot [red]
        table [x expr=\coordindex, y={predictionsRE}]{\CSVPredVSObservedAllExcludedEdgesVarNodes};
        \addlegendentry{Predictions}
        
        \addplot [blue]
        table [x expr=\coordindex, y={true_valuesRE}]{\CSVPredVSObservedAllExcludedEdgesVarNodes};
        \addlegendentry{Labels}
        
        \addplot [black, forget plot]
        table [x expr=\coordindex, y={max_voltagesRE}]{\CSVPredVSObservedAllExcludedEdgesVarNodes};
        
        \addplot [black]
        table [x expr=\coordindex, y={min_voltagesRE}]{\CSVPredVSObservedAllExcludedEdgesVarNodes};
        \addlegendentry{Bounds}

        \nextgroupplot[
        yscale=0.833,
        tick align=outside,
        tick pos=left,
        x grid style={white!69.0196078431373!black},
        xlabel={Bus index ($i$)},
        xmin=-1.45, xmax=30.45,
        xtick style={color=black},
        y grid style={white!69.0196078431373!black},
        ylabel={\(\displaystyle \Im({V}_i)\)},
        ymin=-0.537296454651957, ymax=1.0,
        ytick style={color=black},
        xmajorgrids=true,
        ymajorgrids=true,
        grid style=dashed
        ]
        
        \addplot [red]
        table [x expr=\coordindex, y={predictionsIM}]{\CSVPredVSObservedAllExcludedEdgesVarNodes};
        
        \addplot [blue]
        table [x expr=\coordindex, y={true_valuesIM}]{\CSVPredVSObservedAllExcludedEdgesVarNodes};
        
        \addplot [black, forget plot]
        table [x expr=\coordindex, y={max_voltagesIM}]{\CSVPredVSObservedAllExcludedEdgesVarNodes};
        
        \addplot [black]
        table [x expr=\coordindex, y={min_voltagesIM}]{\CSVPredVSObservedAllExcludedEdgesVarNodes};
        
        \end{groupplot}
        
        \end{tikzpicture}
    \caption{Predictions and labels for one test example with 49 measurement phasors removed.}
    \label{PredictionsPerNode49Excluded}
\end{figure}
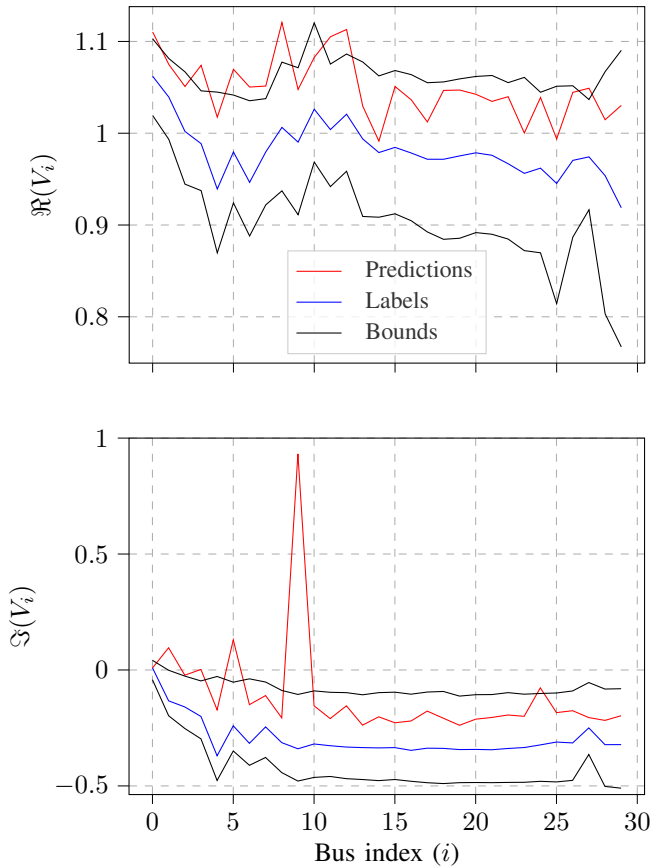

Fig. \ref{PredictionsPerNode2Excluded} and Fig. \ref{PredictionsPerNode49Excluded} depict predictions and labels for the model evaluated on the same test example with 2 and 49 measurement phasors excluded, respectively. We can observe that GNN predictions are a decent fit for many of the node labels when small fractions of all PMU measurements are not delivered to the state estimator, demonstrating robustness in these unobservable scenarios. In the unlikely scenario of only one measurement phasor left in the power system, the proposed model generates predictions within  the scope of the depicted bounds for most of the variable nodes, although GNN inputs, in this case, differ significantly from the samples on which the proposed model was trained.

\begin{figure}[htbp]
    \centering
    \pgfplotstableread[col sep = comma,]{data/PredictionsPerNode2PMUsExcluded.csv}\CSVPredVSObservdTwoPMUExcludedEdgesVarNodes
    
        \begin{tikzpicture}
        
        \begin{groupplot}[group style={group size=1 by 2}, ]
        \nextgroupplot[
        yscale=0.833,
        legend cell align={left},
        legend columns=1,
        legend style={
          fill opacity=0.8,
          font=\small,
          draw opacity=1,
          text opacity=1,
          at={(0.19,0.760)},
          anchor=south,
          draw=white!80!black,
          nodes={scale=0.98, transform shape}
        },
        scaled x ticks=manual:{}{\pgfmathparse{#1}},
        tick align=outside,
        tick pos=left,
        x grid style={white!69.0196078431373!black},
        xmin=-1.45, xmax=30.45,
        xtick style={color=black},
        xticklabels={},
        y grid style={white!69.0196078431373!black},
        ylabel={\(\displaystyle \Re({V}_i)\)},
        ymin=0.749538275942936, ymax=1.13792501106016,
        ytick style={color=black}
        ]
        \addplot [red]
        table [x expr=\coordindex, y={predictionsRE}]{\CSVPredVSObservdTwoPMUExcludedEdgesVarNodes};
        \addlegendentry{Predictions}
        
        \addplot [blue]
        table [x expr=\coordindex, y={true_valuesRE}]{\CSVPredVSObservdTwoPMUExcludedEdgesVarNodes};
        \addlegendentry{Labels}
        
        \addplot [black, forget plot]
        table [x expr=\coordindex, y={max_voltagesRE}]{\CSVPredVSObservdTwoPMUExcludedEdgesVarNodes};
        
        \addplot [black]
        table [x expr=\coordindex, y={min_voltagesRE}]{\CSVPredVSObservdTwoPMUExcludedEdgesVarNodes};
        \addlegendentry{Bounds}
        
        \addplot [very thin, black, dashed, forget plot]
        table {%
        14 -0.703416573701494
        14 1.31495756705672
        };
        \addplot [very thin, black, dashed, forget plot]
        table {%
        44 -0.703416573701494
        44 1.31495756705672
        };
        \addplot [very thin, black, dashed, forget plot]
        table {%
        11 -0.703416573701494
        11 1.31495756705672
        };
        \addplot [very thin, black, dashed, forget plot]
        table {%
        41 -0.703416573701494
        41 1.31495756705672
        };
        \addplot [very thin, black, dashed, forget plot]
        table {%
        13 -0.703416573701494
        13 1.31495756705672
        };
        \addplot [very thin, black, dashed, forget plot]
        table {%
        43 -0.703416573701494
        43 1.31495756705672
        };
        \addplot [very thin, black, dashed, forget plot]
        table {%
        22 -0.703416573701494
        22 1.31495756705672
        };
        \addplot [very thin, black, dashed, forget plot]
        table {%
        52 -0.703416573701494
        52 1.31495756705672
        };
        \addplot [very thin, black, dashed, forget plot]
        table {%
        17 -0.703416573701494
        17 1.31495756705672
        };
        \addplot [very thin, black, dashed, forget plot]
        table {%
        47 -0.703416573701494
        47 1.31495756705672
        };
        \addplot [very thin, black, dashed, forget plot]
        table {%
        18 -0.703416573701494
        18 1.31495756705672
        };
        \addplot [very thin, black, dashed, forget plot]
        table {%
        48 -0.703416573701494
        48 1.31495756705672
        };

        \nextgroupplot[
        yscale=0.833,
        tick align=outside,
        tick pos=left,
        x grid style={white!69.0196078431373!black},
        xlabel={Bus index ($i$)},
        xmin=-1.45, xmax=30.45,
        xtick style={color=black},
        y grid style={white!69.0196078431373!black},
        ylabel={\(\displaystyle \Im({V}_i)\)},
        ymin=-0.537296454651957, ymax=0.0692334804622149,
        ytick style={color=black}
        ]
        
        \addplot [red]
        table [x expr=\coordindex, y={predictionsIM}]{\CSVPredVSObservdTwoPMUExcludedEdgesVarNodes};
        
        \addplot [blue]
        table [x expr=\coordindex, y={true_valuesIM}]{\CSVPredVSObservdTwoPMUExcludedEdgesVarNodes};
        
        \addplot [black, forget plot]
        table [x expr=\coordindex, y={max_voltagesIM}]{\CSVPredVSObservdTwoPMUExcludedEdgesVarNodes};
        
        \addplot [black]
        table [x expr=\coordindex, y={min_voltagesIM}]{\CSVPredVSObservdTwoPMUExcludedEdgesVarNodes};
        
        \addplot [very thin, black, dashed, forget plot]
        table {%
        14 -0.703416573701494
        14 1.31495756705672
        };
        \addplot [very thin, black, dashed, forget plot]
        table {%
        44 -0.703416573701494
        44 1.31495756705672
        };
        \addplot [very thin, black, dashed, forget plot]
        table {%
        11 -0.703416573701494
        11 1.31495756705672
        };
        \addplot [very thin, black, dashed, forget plot]
        table {%
        41 -0.703416573701494
        41 1.31495756705672
        };
        \addplot [very thin, black, dashed, forget plot]
        table {%
        13 -0.703416573701494
        13 1.31495756705672
        };
        \addplot [very thin, black, dashed, forget plot]
        table {%
        43 -0.703416573701494
        43 1.31495756705672
        };
        \addplot [very thin, black, dashed, forget plot]
        table {%
        22 -0.703416573701494
        22 1.31495756705672
        };
        \addplot [very thin, black, dashed, forget plot]
        table {%
        52 -0.703416573701494
        52 1.31495756705672
        };
        \addplot [very thin, black, dashed, forget plot]
        table {%
        17 -0.703416573701494
        17 1.31495756705672
        };
        \addplot [very thin, black, dashed, forget plot]
        table {%
        47 -0.703416573701494
        47 1.31495756705672
        };
        \addplot [very thin, black, dashed, forget plot]
        table {%
        18 -0.703416573701494
        18 1.31495756705672
        };
        \addplot [very thin, black, dashed, forget plot]
        table {%
        48 -0.703416573701494
        48 1.31495756705672
        };
        
        \end{groupplot}
        
        \end{tikzpicture}

    \caption{Predictions and labels for one test example with phasors from two neighboring PMUs removed.}
    \label{PredictionsPerNode2PMUsExcluded}
\end{figure}
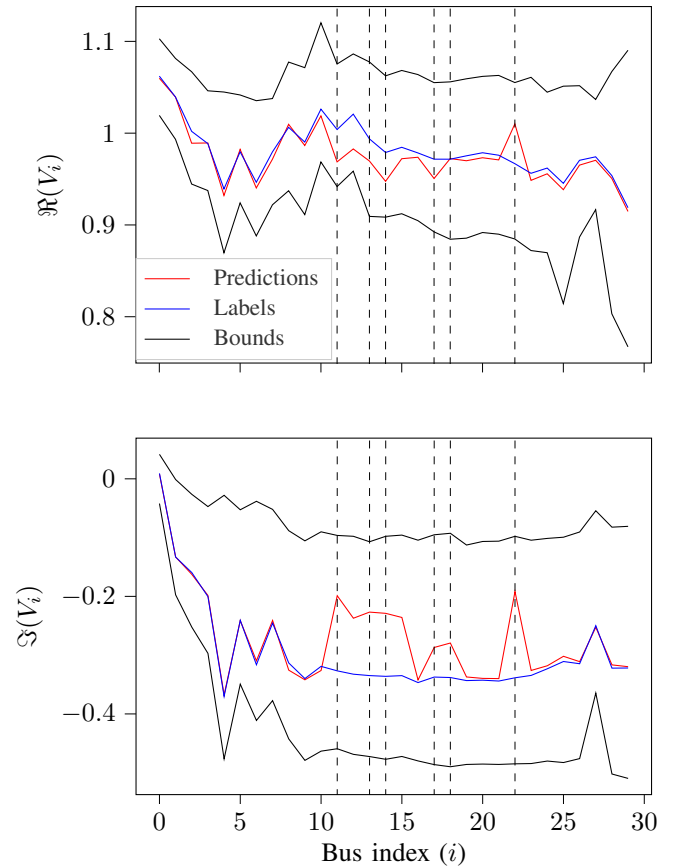

To further analyze the robustness of the proposed model, we observe the predictions for the scenario where 2 neighboring PMUs fail to deliver measurements to the state estimator, hence all of the 8 measurement phasors associated to the removed PMUs are excluded from the GNN inputs. Average Pearson’s Correlation and MSE for the test set of 100 samples created by removing these measurements from the original test set used throughout this section are equal to $0.9985532$ and $1.3815\times 10^{-3}$. Predictions and labels per variable node index for one sample are shown in Fig \ref{PredictionsPerNode2PMUsExcluded}, in which vertical dashed lines indicate indices of the variable nodes within 1-hop neighbourhood of the removed PMUs. We stress that significant deviations from the labels occur for the neighboring nodes only, with no greater effect on the predictions for the remaining nodes, making the proposed model a suitable tool to be used in PMU failure scenarios.

\subsection{Sample efficiency analysis}

To demonstrate how sample efficient is the proposed method, it is trained on three training sets with sizes of 100, 1000, and 10000, for which validation set losses converged after 10000, 1000, and 100 epochs respectively. All three trained models are tested on the same test set of size 100, which contains no excluded measurements. The average Pearson's Correlation and MSE between the labels and the predicted bus voltages over the entire test set are shown in Table~\ref{tbl_sample_efficiency}. It can be seen that models trained with 1000 and 10000 samples produce results of similar quality during the test phase, whereas there is a more noticeable deterioration of results for the model trained with 100 samples. We assume that further exponential increase in training set size would not be followed by significant improvements in the displayed performance indicators.

\begin{table}[htbp]
\caption{Test set results for different training set sizes.}
\begin{center}
    \begin{tabular}{ | c | c | c | }
        \hline
        \textbf{Training set size} & \textbf{Pearson’s Correlation} & \textbf{MSE} \\
        \hline
        100 & 0.9999642 & $1.5329\times 10^{-4}$ \\
        \hline
        1000 & 0.9999909 & $1.4891\times 10^{-5}$ \\
        \hline
        10000 & 0.9999948 & $1.0512\times 10^{-5}$ \\
        \hline
    \end{tabular}
\label{tbl_sample_efficiency}
\end{center}
\end{table}

\section{Conclusions}
In this paper, we present a study on possibilities to use GNNs as fast linear SE with PMUs solvers. We propose a model with graph attention network based architecture, which operates on heterogeneous graphs, containing variable nodes that output the predictions based on the WLS SE solutions, and factor nodes, which take PMU voltage and current measurements and variances as inputs and propagate them in the local neighbourhood. Evaluating the trained model on the unseen data samples confirms that the proposed GNN approach can be used as a very accurate approximator of the WLS SE solutions. By testing the model on scenarios in which individual phasor measurements, or the whole PMUs fail to deliver measurement data to the proposed SE solver, we have found that adding variable-to-variable node connections in the training and test graphs significantly improves the predictions in the cases when the system of equations defining the SE problem becomes underdetermined. 


\bibliographystyle{IEEEtran}
\bibliography{bibliography}

\end{document}